\pdfoutput=1

\documentclass[11pt]{article}

\usepackage{naacl2021}

\usepackage{times}
\usepackage{latexsym}

\usepackage[T1]{fontenc}

\usepackage[utf8]{inputenc}

\usepackage{microtype}

\usepackage{url}
\usepackage{graphicx}
\usepackage{xspace}
\usepackage{xcolor}
\usepackage{booktabs}
\usepackage{array}
\usepackage[inline]{enumitem}
\usepackage{amsmath}
\usepackage{comment}
\usepackage{multirow}
\usepackage{amssymb}

\newcommand{\papertitle}{Semantic Frame Forecast}

\newcommand{\Task}{Semantic Frame Forecast\xspace}
\newcommand{\task}{semantic frame forecast\xspace}

\newcommand{\kenneth}[1]{}
\newcommand{\cy}[1]{}
\title{\papertitle}

\author{Chieh-Yang Huang and Ting-Hao (Kenneth) Huang\\
  Pennsylvania State University, University Park, PA 16802, USA\\
  \texttt{\{chiehyang,~txh710\}@psu.edu} \\}

\begin{document}
\maketitle
\begin{abstract}

This paper introduces \textbf{\task}, a task that predicts the semantic frames that will occur in the next 10, 100, or even 1,000 sentences in a running story.
Prior work focused on predicting the immediate future of a story, such as one to a few sentences ahead.
However, when novelists write long stories, generating a few sentences is not enough to help them gain high-level insight to develop the follow-up story.
In this paper, we formulate a long story as a sequence of ``story blocks,'' where each block contains a fixed number of sentences ({\em e.g.,} 10, 100, or 200).
This formulation allows us to predict the follow-up story arc beyond the scope of a few sentences.
We represent a story block using the term frequencies (TF) of \textbf{semantic frames} in it, normalized by each frame's inverse document frequency (IDF).
We conduct \task experiments on 4,794 books from the Bookcorpus and 7,962 scientific abstracts from CODA-19, with block sizes ranging from 5 to 1,000 sentences.
The results show that automated models can forecast the follow-up story blocks better than the random, prior, and replay baselines, indicating the task's feasibility.
We also learn that the models using the frame representation as features outperform all the existing approaches when the block size is over 150 sentences.
The human evaluation also shows that the proposed frame representation, when visualized as word clouds, is comprehensible, representative, and specific to humans.
Our code is available at: \url{https://github.com/appleternity/FrameForecasting}.

\end{abstract}

\section{Introduction}
\label{sec:intro}

\begin{figure}[t]
    \centering
    \includegraphics[width=1\columnwidth]{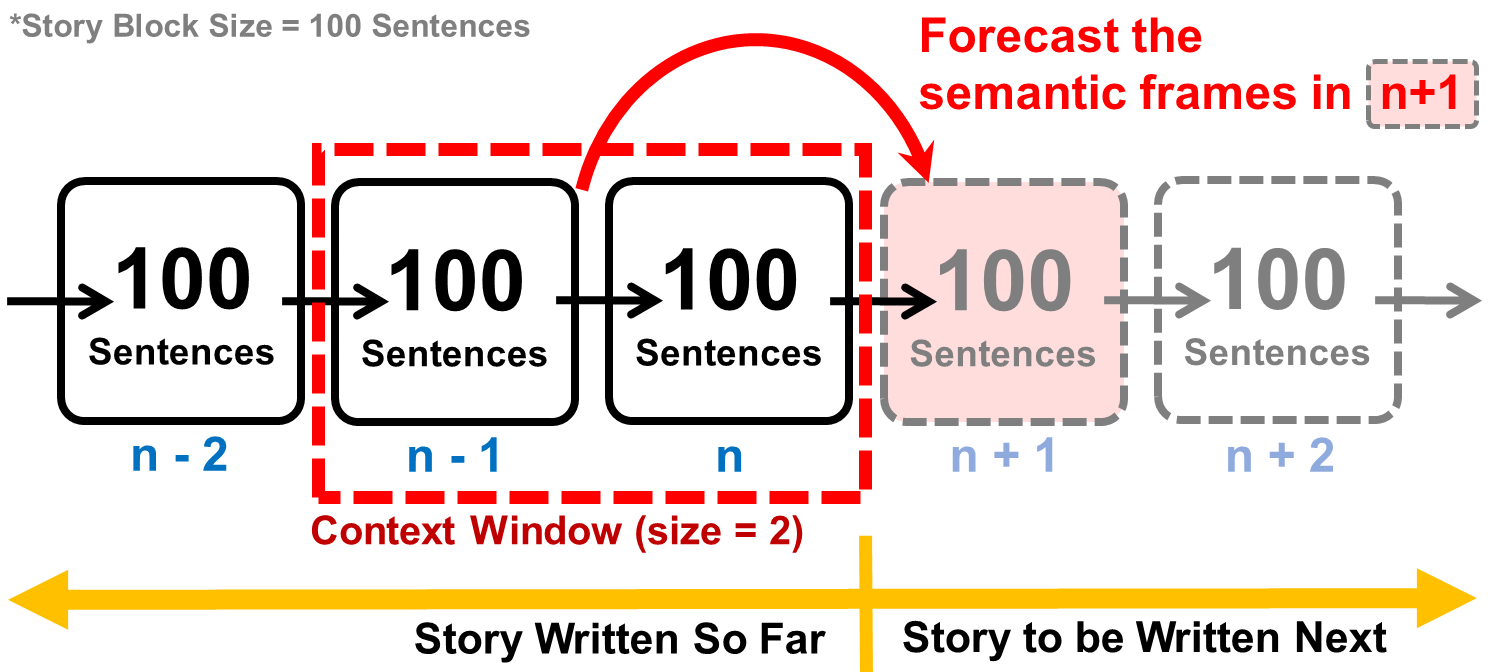}
    \caption{The \task is a task that predicts the semantic frames that will occur in the next part of a story based on the texts written so far.}
    \label{fig:example-ff}
\end{figure}

Writing a good novel is hard.
Creative writers can get stuck in the middle of their drafts and struggle to develop follow-up scenes.
Writing support systems, such as Heteroglossia~\cite{Heteroglossia}, generate paragraphs or ideas to help writers figure out the next part of the ongoing story.
However, little literature focuses on plot prediction for \textbf{long stories}.
Much prior work focused on predicting the immediate future of a story, {\em i.e.,} one to a few sentences later.
For example, the Creative Help system used a recurrent neural network model to generate the next sentence to support writing~\cite{roemmele2015creative};
the Scheherazade system uses crowdsourcing and artificial intelligence techniques to interactively construct the narrative sentence by sentence~\cite{li2015scheherazade}; Clark et al.~\shortcite{Clark:2018:CWM:3172944.3172983} study machine-in-the-loop story writing where the machine constantly generates a suggestion for the next sentence to stimulate writers;
and Metaphoria~\cite{10.1145/3290605.3300526} generates metaphors, an even smaller unit, to inspire writers based on an input word by searching relations and ranking distances on ConceptNet~\cite{liu2004conceptnet}.

Generating a coherent story across multiple sentences is challenging, even with cutting-edge pretrained models~\cite{see-etal-2019-massively}. 
To generate coherent stories, researchers often first generate a high-level representation of the story plots and then use it as a guide to generate a full story.
For example, Martin et al.~\shortcite{martin2018event} propose an event representation that uses an \textit{SVO} tuple to generate story plots; 
Plan-and-write~\cite{yao2019plan} uses the RAKE algorithm~\cite{RAKE} to extract the keyword in each sentence to form a storyline and treat it as an intermediate representation; 
Fan et al.~\shortcite{fan-etal-2019-strategies} use predicate-argument pairs annotated by semantic role labelers to model the structure of stories; 
and Zhang et al.~\shortcite{zhang2020visual} take words with a certain part-of-speech tag as anchors and show that using anchors as the intermediate representation can improve the story quality.
However, these projects all focused on short stories:
The event representation is developed on a Wikipedia movie plot summary dataset~\cite{bamman2013learning}, where a summary has an average of 14.52 sentences;
Plan-and-write uses the ROCStories dataset~\cite{mostafazadeh-etal-2016-corpus}, where each story has only 5 sentences;
Fan {\em et al.} test their algorithm on the WritingPrompts dataset~\cite{fan-etal-2018-hierarchical}, where stories have 734 words (around 42 sentences) on average;
and Zhange {\em et al.}'s anchor representation is developed on the VIST dataset~\cite{huang-etal-2016-visual}, where a story has 5 sentences.

All the existing intermediate representations are generated on a sentence basis, meaning that the length of the representations increases along with the story length.
That is, when applying these representations to \textbf{novels that usually have more than 50,000 words} (as defined by the National Novel Writing Month~\cite{wikipedia_2020}), it is not likely that such representations can still work.
We thus introduce a new \textbf{Frame Representation} that compiles semantic frames into a fixed-length TF-IDF vector and a \textbf{\Task} task that aims to predict the next frame representation using the information in the current story block (see Figure~\ref{fig:example-ff}).
Two different datasets are built to examine the effectiveness of the proposed frame representation: one from Bookcorpus~\cite{Zhu_2015_ICCV}, a fiction dataset; and one from CODA-19~\cite{huang2020coda}, a scientific abstract dataset.
We establish several baselines and test them on different story block sizes, up to 1,000 sentences.
The result shows that the proposed frame representation successfully captures the story plot information and helps the \task task, especially for story blocks with more than 150 sentences.
To enable humans to perceive and comprehend frame representations, we further propose a process that visualizes a vector-based frame representation as word clouds.
Human evaluations show that 
word clouds represent a story block with reasonable specificity, 
and
our proposed model produces word clouds that are more representative than that of BERT.

\section{Related Work}

\paragraph{Automated Story Generation.}
Classic story generation focuses on generating logically coherent stories, plot planning~\cite{riedl2010narrative,li2013story}, and case-based reasoning~\cite{gervas2004story}.
Recently, several neural story generation models have been proposed~\cite{peng-etal-2018-towards,fan-etal-2018-hierarchical}, even including massive pretrained models~\cite{radford2019language,keskar2019ctrl}.
However, researchers realize that word-by-word generation models cannot efficiently model the long dependency across sentences~\cite{see-etal-2019-massively}.
Models using intermediate representations as guidance to generate stories are then proposed~\cite{yao2019plan,martin2018event,ammanabrolu2020story,fan-etal-2019-strategies,zhang2020visual}. 
These works are developed toward short stories and thus are insufficient for modeling novels (See Section~\ref{sec:intro}).

\paragraph{Automated Story Understanding.}
Story understanding is a longstanding goal of AI~\cite{roemmele2018encoder}.
Several tests were proposed to evaluate AI models' ability to reason the event sequence in a story.
Roemmele {\em et al.}~\shortcite{roemmele2011choice} proposed the Choice of Plausible Alternatives (COPA) task, focusing on commonsense knowledge related to identifying causal relations between sequences.
Mostafazadeh {\em et al.}~\shortcite{mostafazadeh-etal-2016-corpus} proposed the Story Cloze Test, in which the model is required to select which of two given sentences best completes a particular story. Ippolito {\em et al.}~\shortcite{ippolito2019unsupervised} proposed the Story Infilling task, which aims to generate the middle span of a story that is coherent with the foregoing context and will reasonably lead to the subsequent plots.
Under the broader umbrella of story understanding, some prior work aimed to predict the next event in a story~\cite{granroth2016happens} or to identify the right follow-up line in dialogues~~\cite{lowe2016evaluation}.

\section{\Task}

\begin{figure}[t]
    \centering
    \includegraphics[width=1\columnwidth]{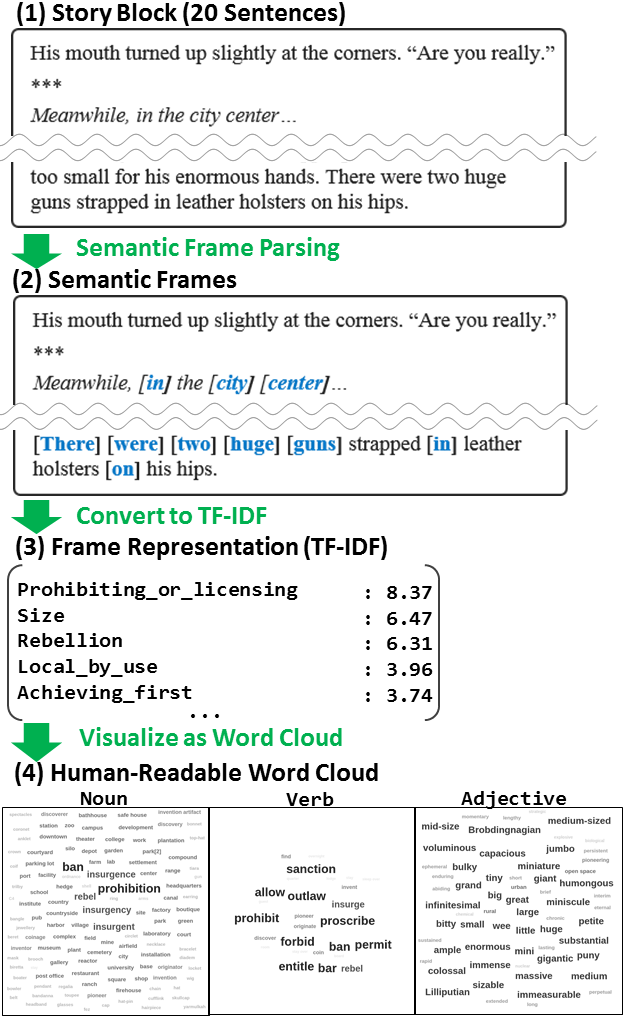}
    \caption{The steps to generate the frame representation for story blocks. The human-readable word clouds are generated to illustrate the conceptual meaning of the frame representation.}
    \label{fig:story-block}
\end{figure}

As shown in Figure~\ref{fig:example-ff}, we formulate a long story as a sequence of fixed-length story blocks.
Each story block (Figure~\ref{fig:story-block} (1)) has a set of semantic frames (Figure~\ref{fig:story-block} (2))~\cite{Baker:1998:BFP:980845.980860}. 
We convert a story block into the \textbf{Frame Representation} (Figure~\ref{fig:story-block} (3)), a TF-IDF vector over semantic frames, by computing the term frequency in that story block and the inverse document frequency over all the story blocks in the corpus.
FrameNet~\cite{Baker:1998:BFP:980845.980860} defined a total of 1,221 different semantic frames, so the generated TF-IDF has 1,221 dimensions.
The \textbf{\Task} is then defined as a task to predict the frame representation of the \textit{n+1}-th story block using the foregoing content, namely the \textit{n}-th story block.


\paragraph{Evaluation Metric.}
We use Cosine Similarity between the predicted vector and the gold-standard vector (complied from the human-written story block) for evaluation.
Many other metrics, such as Mean-Squared Error (MSE), also exist to measure the distance between two vectors.



\section{Data}

\begin{table*}[t]
\centering \small
    \addtolength{\tabcolsep}{-0.08cm}
    \begin{tabular}{lrrrrrrrrrr}
    \toprule \midrule
    \textbf{Block Size} & \multicolumn{1}{c}{\textbf{5}} & \multicolumn{1}{c}{\textbf{10}} & \multicolumn{1}{c}{\textbf{20}} & \multicolumn{1}{c}{\textbf{50}} & \multicolumn{1}{c}{\textbf{100}} & \multicolumn{1}{c}{\textbf{150}} & \multicolumn{1}{c}{\textbf{200}} & \multicolumn{1}{c}{\textbf{300}} & \multicolumn{1}{c}{\textbf{500}} & \multicolumn{1}{c}{\textbf{1000}} \\ \midrule
    \# Words Mean & 71.7 & 143.5 & 286.9 & 717.2 & 1433.9 & 2149.8 & 2865.3 & 4293.7 & 7142.5 & 14212.3 \\
    \# Frames Mean & 17.5 & 35.0 & 69.9 & 174.5 & 348.6 & 522.1 & 695.4 & 1040.7 & 1727.3 & 3417.2 \\
    \# Events Mean & 10.0 & 20.0 & 39.9 & 99.8 & 199.4 & 298.9 & 398.2 & 596.4 & 991.2 & 1967.1 \\
    \# Train & 3,744,948 & 1,869,947 & 932,464 & 369,941 & 182,479 & 119,967 & 88,720 & 57,455 & 32,469 & 13,749 \\
    \# Valid & 574,840 & 287,054 & 143,166 & 56,838 & 28,073 & 18,466 & 13,672 & 8,881 & 5,035 & 2,166 \\
    \# Test & 1,054,816 & 526,687 & 262,625 & 104,198 & 51,396 & 33,776 & 24,987 & 16,178 & 9,138 & 3,861 \\ \midrule \bottomrule
    \end{tabular}
    \addtolength{\tabcolsep}{+0.08cm}
    \caption{The statistic of Bookcorpus dataset in ten different story block lengths. We use Open-Sesame to parse the semantic frame for each sentence. The \textit{Events} represents the SVO tuples \cite{martin2018event}.}
    \label{tab:dataset-stat-bookcorpus}
\end{table*}

\begin{table}[t]
    \small \centering
    \begin{tabular}{lrrr}
    \toprule \midrule
    \multicolumn{1}{c}{\textbf{Block Size}} & \multicolumn{1}{c}{\textbf{1}} & \multicolumn{1}{c}{\textbf{3}} & \multicolumn{1}{c}{\textbf{5}} \\ \midrule
    \# Words Mean & 26.3 & 77.3 & 124.7 \\
    \# Frames Mean & 6.0 & 17.5 & 27.6 \\
    \# Events Mean & 1.2 & 3.5 & 5.6 \\
    \# Train & 48,489 & 9,858 & 2,739 \\
    \# Valid & 5,615 & 1,146 & 334 \\
    \# Test & 5,238 & 1,047 & 287 \\ \midrule \bottomrule
    \end{tabular}
    \caption{The statistic of CODA-19 dataset in three different story block lengths. We use Open-Sesame to parse the semantic frame for each sentence. The \textit{Events} represents the SVO tuples \cite{martin2018event}.}
    \label{tab:dataset-stat-coda19}
\end{table}




We build the dataset from the existing Bookcorpus dataset~\cite{Zhu_2015_ICCV} and CODA-19 dataset~\cite{huang2020coda}.
This section describes how we preprocess the data, remove undesired content, and build the final dataset.

\paragraph{Bookcorpus Dataset.}

We obtain a total of $15,605$ raw books and their corresponding meta data.
To get high-quality fictional content, we remove books using the following heuristic rules:
\begin{enumerate*}[label=(\roman*)]
    \item short books whose size is less than 10KB;
    \item books that contain HTML code;
    \item books that are in the epub format (an e-book file format);
    \item books that are not in English;
    \item books that are in the ``Non-Fiction'' genre;
    \item books that are in the ``Anthologies'' genre;
    \item books that are in the ``Graphic Novels \& Comics'' genre.
\end{enumerate*}
Since most books contain book information, author information, and some nonfictional content at the beginning and end of the book, we use regular expressions to match the term ``Chapter'' to locate the chapter title.
Only the contents between the first chapter title and the last chapter title are kept.
The last chapter is also removed as there are no certain boundaries to identify the story ending.
Books whose chapter titles are unlocatable are also removed.
After removing all the unqualified books, a total of $4,794$ books were used in our dataset.
We transliterate all non-ASCII characters into ASCII characters using Unidecode (https://pypi.org/project/Unidecode/) to fulfill the requirement of Open-SESAME~\cite{swayamdipta:17}.
Open-SESAME is then used to parse the semantic frames for each sentence.

The books are split into training/validation/test sets following a 70/10/20 split, resulting in $3,357$, $479$, and $958$ books, respectively.
To measure the effect of frame representation for different context lengths, we vary the story block length, using $5$, $10$, $20$, $50$, $100$, $150$, $200$, $300$, $500$, and $1,000$ sentences.
When creating instances, we first split a book into story blocks with the specified length and extract all the consecutive two story blocks as instances when context window size (see Figure~\ref{fig:example-ff}) is set to 1.
The IDF of the semantic frame is then computed over the story blocks using all the training sets.
Combining with the TF value in each story block, we convert story blocks into frame representations.
We use scikit-learn's implementation~\cite{scikit-learn} of TF-IDF but with a slight modification on IDF: Scikit-learn uses $idf(t) = log(\frac{n}{df(t) + 1})$ to compute a smoothing IDF, but we use $idf(t) = log (\frac{n}{df(t)})$.
The detailed statistic information is shown in Table~\ref{tab:dataset-stat-bookcorpus}.

\paragraph{CODA-19 Dataset.}
\cy{Add motivation of choosing CODA-19.}
We envision a broader definition of ``creativity'' in writing and attempt to apply story arc prediction technologies to the domains outside novels, for example, scholarly articles.
As an earlier exploration, we choose to use a smaller set of human-annotated abstracts (CODA-19~\cite{huang2020coda}) rather than machine-extracted full text (CORD-19~\cite{wang2020cord}) in our proof-of-concept study, avoiding formatting issues ({\em e.g.}, reference format, parsing errors) and intensive data cleaning effort.
The original CODA-19 dataset contains $10,966$ human-annotated English abstracts for five different aspects: Background, Purpose, Method, Finding/Contribution, and Other.
We remove sentences that are annotated as ``Other,'' an aspect for sentences that are not directly related to the content ({\em e.g.,} terminology definitions or copyright notices.)
Abstracts that contain Unicode characters are also removed.
A total of $7,962$ abstracts are used in our dataset.
We then use Open-SESAME to parse the semantic frames for each sentence.
We adopt CODA-19's original split, where the training set, validation set, and testing set have $6,509$, $737$, and $716$ abstracts, respectively.
Three different lengths of story block are used: $1$, $3$, and $5$.
We then create instances and compute TF-IDF as described above.
Table~\ref{tab:dataset-stat-coda19} shows the details.

\section{Models}

We implement two naive baselines, an information retrieval baseline, two machine learning baselines, two deep learning baselines, an existing model and a text generation baseline.



\paragraph{Replay Model.}

For each instance, the replay model takes the frame representation in the \textit{n}-th story block as the prediction, {\em i.e.}, the same frames will occur again.

\paragraph{Prior Model.}

The prior model computes the mean of the frame representation over the training set and uses it as the prediction for all the testing instances.

\paragraph{Information Retrieval with Frame Representation.}

For each instance, the information retrieval model searches for the most similar story block in the training set and takes the frame representation from its next story block as the prediction.
In this setting, we adopt the cosine similarity on frame representations to measure the story similarity.
For block size 5 in the Bookcorpus dataset, there are around 3.7 million instances in the training set, which is infeasible to finish.

\paragraph{Random Forest with Frame Representation.}

The foregoing story block's frame representation is used as the feature for prediction.
We use scikit-learn's implementation of Random Forest Regressor~\cite{scikit-learn} with a max depth of 3 and 20 estimators.
For block sizes that have more than one million training instances (5 and 10 in the Bookcorpus dataset), we randomly sample one million instances to train the model.

\paragraph{LGBM with Frame Representation.}

This is the same as the previous setup but trained using the LGBM Regressor model~\cite{ke2017lightgbm} with the max depth 5, the number of leaves 5, and the number of estimators 100.
For block sizes that have more than one million training instances (5 and 10 in the Bookcorpus dataset), we randomly sample one million instances to train the model.

\paragraph{DAE with Frame Representation.}

This is the same as the previous setting but trained with the Denoising Autoencoder architecture~\cite{bengio2013generalized}.
We feed in the foregoing story block's frame representation and output the frame representation for the follow-up story block.
Thirty percent of the input is dropped randomly.
The model is optimized using the cosine distance ($1-cosine\ similarity$).
Both the encoder and decoder are created via five dense layers with a hidden size of 512.
We use a learning rate of 1e-5 and a batch size of 512 and train the model with the early stopping criteria of no improvement for 20 epochs. 
The best model on the validation set is kept for testing.

\paragraph{Event Representation Model (Event-Rep).}

We use Martin {\em et al.}'s event representation~\shortcite{martin2018event} on the foregoing story block as the feature.
An event tuple is defined as $\left \langle s, v, o, m \right \rangle$, where $s$ is the subject, $v$ is the verb, $o$ is the object, and $m$ is the verb modifier.
We extract the dependency relation using the Stanza parser~\cite{qi2020stanza}.
Unlike Martin {\em el al.}'s implementation, where the empty placeholder ${\varnothing}$ only replaces unidentified objects and modifiers, we find that the subjects can also be frequently missing in fiction books. 
For example, in ``\textit{``\textbf{Come} out?'' Zack asked. ``\textbf{Come} out of where?''}''. In both cases here, the verb ``come'' does not have a subject.
In ``\textit{Fine, \textbf{follow} me.}'', ``follow'' has an object but does not have a subject.
Therefore, we allow $s$ to have a ${\varnothing}$ placeholder in our implementation.
All words are stemmed by NLTK~\cite{10.3115/1118108.1118117}.

After extracting the event representation, the sequence of event tuples in the foregoing story block is fed into a five-layer LSTM model~\cite{hochreiter1997long} to predict its follow-up frame representation.
Note that the length of the event tuple sequence changes along with the block size.
We thus set the maximum length of the sequence to the 95th percentile of the length in the training set.
Sequences longer than the maximum length are left-truncated.
The model is trained with a hidden size of 512, a learning rate of 3e-5, a dropout rate of 0.05, and a batch size of 64. 
We optimize the model using the cosine distance and apply the early stopping criteria of no improvement for three epochs.
The best model on the validation set is kept for testing.

\paragraph{BERT.}
We take the pure text in the foregoing story block as the feature and apply the pretrained BERT model~\cite{devlin2018bert}.
BERT has a token length limitation, so we set the maximum length of tokens to 500 for Bookcorpus and 300 for CODA-19.
Sentences with more than 500 tokens are truncated from the left.
We take the \texttt{[CLS]} token representation from the last layer and add a dense layer on top of it to predict the follow-up frame representation.
The model is trained with a learning rate of 1e-5 and a batch size of 32.
We optimize the model using the cosine distance and apply the early stopping when no improvement for five epochs.
The model with the best score on the validation set is kept for testing.

\paragraph{SciBERT (For CODA-19 Only).}
This is the same as the previous setting but is trained using the pretrained SciBERT model~\cite{Beltagy2019SciBERT}.
We only test this approach on the CODA-19 dataset since it is from the scientific domain.

\paragraph{GPT-2 (For Bookcorpus Only).}
\cy{Add GPT-2.}
We also include a text generation model, GPT-2 (gpt2-xl) \cite{radford2019language} with block sizes of 5, 10, 20, and 50. 
Since GPT-2 is computationally expensive, we conduct the experiment on a subset of the dataset, where 1,000 instances are randomly selected. 
We feed the text in the latest story block (n) into GPT-2 and generate 70, 150, 300, and 700 words for block sizes 5, 10, 20, and 50, respectively 
(5 sentences $\approx$ 70 words; 10 sentences $\approx$ 150 words in Bookcourpus, etc). 
For stories that exceed the GPT-2's word limit, we truncate the text from the left. 
Stories with block size larger than 100 would have more than 1400 words which by itself exceed the GPT-2's word limit.
Generated stories are then parsed by Open-SESAME to extract the semantic frames and turned into frame representations as the predictions.

\section{Experimental Results and Analysis}
\label{sec:results}
\begin{table*}[t]
\small \center
    \begin{tabular}{llcccccccccc}
    \toprule \midrule
    \multirow{2}{*}{\textbf{Feature}} & \multirow{2}{*}{\textbf{Model}} & \multicolumn{10}{c}{\textbf{Block Size}} \\
    &  & \textbf{5} & \textbf{10} & \textbf{20} & \textbf{50} & \textbf{100} & \textbf{150} & \textbf{200} & \textbf{300} & \textbf{500} & \textbf{1000} \\ \hline
    - & Replay Baseline & .0654 & .0915 & .1237 & .1737 & .2163 & .2448 & .2665 & .3000 & .3462 & .4155 \\
    - & Prior Baseline & .2029 & .2435 & .2857 & .3389 & .3754 & .3962 & .4105 & .4302 & .4528 & .4776 \\ \hline
    Frame & IR Baseline & - & .0631 & .0851 & .1290 & .1841 & .2085 & .2262 & .2536 & .2859 & .3321 \\
    Frame & Random Forest & .2037 & .2448 & .2881 & .3427 & .3807 & .4025 & .4184 & .4402 & .4659 & .4966 \\ 
    Frame & LGBM & .2072 & .2506 & .2967 & .3564 & .3995 & \textbf{.4255} & \textbf{.4441} & \textbf{.4711} & \textbf{.5048} & \textbf{.5510} \\ 
    Frame & DAE & .2082 & .2515 & .2966 & .3547 & .3976 & .4223 & .4400 & .4598 & .4898 & .5280 \\ \hline
    Event & Event-Rep & .2111  & .2541  & .2994  & .3532  & .3929  & .4126  & .4280  & .4453  & .4626  & .4792 \\
    Text & BERT & \textbf{.2172} & \textbf{.2611} & \textbf{.3073} & \textbf{.3637} & \textbf{.4012} & .4229 & .4371 & .4559 & .4779 & .5057 \\
    Text & GPT-2 & .0519 & .0739 & .0990 & .1402 & - & - & - & - & - & - \\ \hline
    & DELTA & .0142 & .0176 & .0216 & .0249 & .0257 & .0293 & .0336 & .0409 & .0520 & .0734 \\
    \midrule \bottomrule
    \end{tabular}
    \caption{Baseline result for Bookcorpus dataset. BERT and Event-Rep work better in smaller block sizes, while models using frame representation perform better in larger block sizes. DELTA represents the difference between the best model and the prior baseline --- an extremely simple but strong baseline --- in that specific block size. The small value of DELTA shows that \task is challenging yet possible.}
    \label{tab:baseline-bookcorpus}
\end{table*}

\begin{table}[t]
\small \center
    \begin{tabular}{llccc}
    \toprule \midrule
    \multirow{2}{*}{\textbf{Feature}} & \multirow{2}{*}{\textbf{Model}} & \multicolumn{3}{c}{\textbf{Block Size}} \\
     & & \textbf{1} & \textbf{3} & \textbf{5} \\ \hline
    - & Replay Baseline & .0524 & .0971 & .1363 \\
    - & Prior Baseline & .1573 & .2067 & .2288 \\ \hline
    Frame & IR Baseline & .0315 & .0601 & .0752 \\
    Frame & Random Forest & .1581 & .2081 & .2278 \\
    Frame & LGBM & .1561 & .2024 & .2094 \\ 
    Frame & DAE & .1611 & .2155 & \textbf{.2380} \\ \hline
    Event & Event-Rep & .1595 & .2118 & .2332 \\ 
    Text & BERT & .1660 & .2202 & .2353 \\
    Text & SciBERT & \textbf{.1675} & \textbf{.2219} & .2339 \\ \hline
    & DELTA & .0102 & .0152 & .0092 \\
    \midrule \bottomrule
    \end{tabular}
    \caption{Baseline result for CODA-19 dataset. SciBERT performs the best in block size 1 and 3. Using the frame representation as the feature, DAE performs the best for block size 5. DELTA shows the difference between the best model and the prior baseline in that specific block size. The small value of DELTA shows that \task is challenging yet possible.}
    \label{tab:baseline-coda19}
\end{table}



Table~\ref{tab:baseline-bookcorpus} and Table~\ref{tab:baseline-coda19} show the experimental results.
In this section, we summarize the main findings.

\paragraph{Predicting forthcoming semantic frames is remarkably challenging yet possible.}
Machine-learning models outperform the two naive baselines for different story lengths.
In the Bookcorpus dataset, BERT performs the best for story blocks under 100 sentences, while LGBM performs the best for story blocks over 150 sentences.
In the CODA-19 dataset, SciBERT performs the best for block sizes of 1 and 3, while DAE performs the best for a block size of 5.
While the task is very challenging, these results shed light on the \task task. 
However, the improvement is not big, as shown in the DELTA row, suggesting that \task requires more investigation and understanding.



\paragraph{``Prior'' is a robust and strong baseline.}
In both the Bookcorpus dataset and the CODA-19 dataset, the prior baseline is strong.
As the story gets longer, the performance also increases. 
This suggests that when the story block gets bigger, more and more frames will constantly occur.

\begin{figure}[h]
    \centering
    \includegraphics[width=1.0\linewidth]{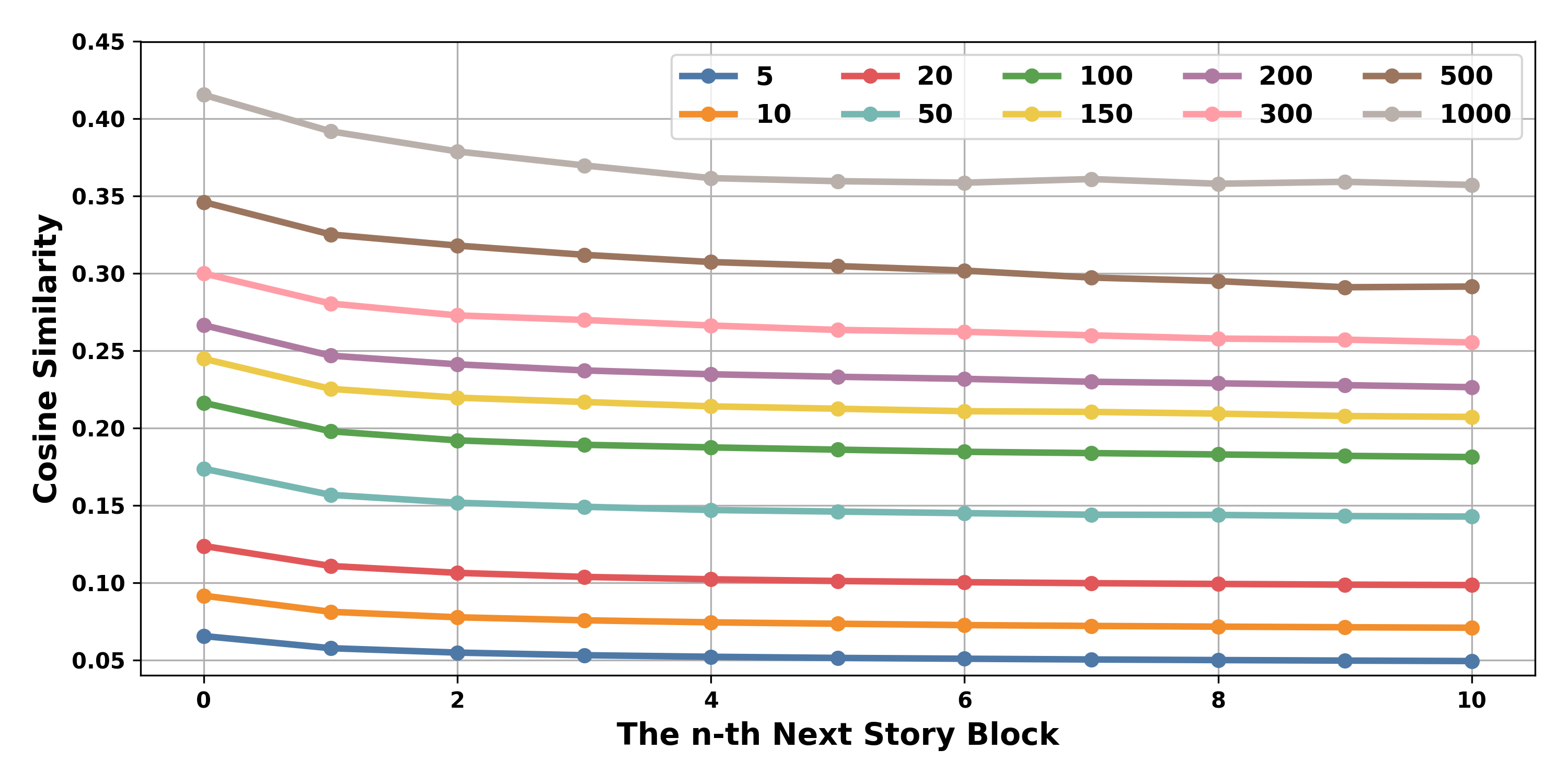}
    \caption{Using the replay baseline to predict the \textit{n+i}-th story block from the \textit{n}-th story block (story block size = 5, 10, $\cdots$, 1000.) Things that happen in the current story block are more likely to happen again shortly.}
    \label{fig:skip-exp}
\end{figure}

\paragraph{Replay baseline shows the relation of consecutive story blocks.}
The replay baseline assumes that the events that happen now will likely happen again shortly.
The results in Table~\ref{tab:baseline-bookcorpus} and Table~\ref{tab:baseline-coda19} partially confirm this assumption.
To understand more about the assumption, we use the replay baseline to predict the \textit{n+i}-th story block from the \textit{n}-th story block in the Bookcorpus dataset.
Figure~\ref{fig:skip-exp} shows the results. 
We can see that things that happen now will be more likely to happen in the near future compared to story blocks farther from the current one.

\paragraph{Event-Rep works better in short stories.}
In the Bookcorpus dataset, event representation works better than the frame representation in small block sizes (5, 10, and 20). 
However, starting from a block size of 50, the model cannot perform as well as the other models. 
We thus conclude that event representation works better in short stories.
The main reason is that event representations are generated on a sentence-by-sentence basis and will create overwhelming information on long stories.
The existing intermediate representations (see Section~\ref{sec:intro}) are mostly generated from sentences and will likely have the same issue as the event representation.
Compared to the existing works, the proposed frame representation encodes a story block, no matter how long it is, into a fixed-length vector and therefore performs better on longer stories.


\paragraph{BERT performs very well in short stories.}
The results of BERT and SciBERT in Table~\ref{tab:baseline-bookcorpus} and Table~\ref{tab:baseline-coda19} show that textual information is helpful in predicting story blocks.
BERT performs better when the block size is under 100 in the Bookcorpus dataset and below 3 in CODA-19.
However, handling long texts remain challenging for BERT, as its computational complexity scales with the square of the token length.
Researchers started reducing the computation complexity for transformer-based models to allow modeling on long texts such as Linformer~\cite{wang2020linformer}, Longformer~\cite{Beltagy2020Longformer}, Reformer~\cite{kitaev2020reformer}, and BigBird~\cite{zaheer2020big}.
However, these models still require a lot of computation power and are not yet ready for general use.

\paragraph{The good performance does not merely come from the number of instances.}
Deep learning methods often require more instances for training.
To show that the result in Table~\ref{tab:baseline-bookcorpus} is not mainly caused by the number of instances, we conduct the same experiment in Bookcorpus dataset using $88,720$ training instances for block sizes ranging from 5 to 200.
Table~\ref{tab:downsample-exp-bookcorpus} shows the results.
The performance is affected, but the conclusions we make above still stand, showing that the number of instances is not the main factor for our observations.
Meanwhile, we find that BERT is affected more than LGBM.
In Table~\ref{tab:downsample-exp-bookcorpus} the performance of BERT drops by $-0.0092$ to $-0.0051$ compared to Table~\ref{tab:baseline-bookcorpus}, but LGBM only drops $-0.0039$ to $-0.0007$.
Although this suggests that the number of instances can cause the difference, it also shows that the 
frame representation can be used with fewer instances.

\begin{table}[t]
    \small \center
    \addtolength{\tabcolsep}{-0.15cm}
    \scalebox{0.9}{
        \begin{tabular}{llccccccc}
        \toprule \midrule
        \multirow{2}{*}{\textbf{Feature}} & \multirow{2}{*}{\textbf{Model}} & \multicolumn{7}{c}{\textbf{Block Size}} \\
        & & \textbf{5} & \textbf{10} & \textbf{20} & \textbf{50} & \textbf{100} & \textbf{150} & \textbf{200} \\ \hline
        - & Prior & .2029 & .2435 & .2856 & .3388 & .3754 & .3962 & .4105 \\ \hline
        Frame & IR & .0401 & .0615 & .0900 & .1368 & .1775 & .2051 & .2262 \\
        Frame & RF & .2030 & .2440 & .2871 & .3418 & .3801 & .4025 & .4184 \\
        Frame & LGBM & .2033 & .2472 & .2935 & .3540 & \textbf{.3980} & \textbf{.4248} & \textbf{.4441} \\
        Frame & DAE & .2058 & .2482 & .2929 & .3507 & .3926 & .4178 & .4400 \\ \hline
        Event & Event-Rep & .2046 & .2470 & .2905 & .3454 & .3799 & .4069 & .4171 \\
        Text & BERT & \textbf{.2088} & \textbf{.2529} & \textbf{.2981} & \textbf{.3550} & .3949 & .4178 & .4371 \\
        \midrule \bottomrule
        \end{tabular}
    }
    \addtolength{\tabcolsep}{+0.15cm}
    \caption{Result of the downsampling experiment. Although all the performance drops, the observations we find are still true. Therefore, the conclusions are not merely caused by the effect of data size.}
    \label{tab:downsample-exp-bookcorpus}
\end{table}

\paragraph{GPT-2 is not effective.}
\cy{add GPT-2 finding.}
GPT-2 is not effective in predicting the story flow even though it can generate reasonable sentences. 
Even the naive Replay baseline outperforms the GPT-2 baseline in predicting the story block.
We hypothesize that GPT-2 is not good at maintaining the coherence among sentences or events, especially in the creative writing domain.
Similar phenomenons are also observed by others and used to motivate the need for guided generation models or progressive generation models~\cite{wang2020narrative,tan2020progressive}.

\begin{table}[]
    \centering \small
    \addtolength{\tabcolsep}{-0.03cm}
    \begin{tabular}{cllccc}
        \toprule \hline
        \multirow{2}{*}{\textbf{window}} & \multirow{2}{*}{\textbf{Feature}} & \multirow{2}{*}{\textbf{Model}} & \multicolumn{3}{c}{\textbf{Block Size}} \\
                                    & &  & \textbf{20} & \textbf{50} & \textbf{100} \\ \hline
        \multirow{2}{*}{\textbf{2}} & Frame & LGBM        & .2989 & .3590 & \textbf{.4029}\\
        & Text  & BERT                                    & \textbf{.3081} & \textbf{.3625} & .4002 \\ \hline
        \multirow{2}{*}{\textbf{5}} & Frame & LGBM        & .2989 & .3617 & \textbf{.4065}\\
        & Text  & BERT                                    & \textbf{.3082} & \textbf{.3618} & .3985 \\ 
        \hline \bottomrule
    \end{tabular}
    \addtolength{\tabcolsep}{+0.03cm}
    \caption{Results of using 2 or 5 foregoing story blocks to predict the \textit{n+1}-th story block. LGBM improves further when using more context but BERT fails to model the longer context, and its performance even gets hurt.}
    \label{tab:history-exp}
\end{table}

\subsection{Using a Larger Context Window}
This paper focuses on using 1 story block to forecast the next one, {\em i.e.,} window size = 1 (see Figure~\ref{fig:example-ff}.)
As a proof of concept, we use 2 and 5 blocks (window size = 2 and 5) for prediction, respectively.
We use two models:
LGBM with frame representation, and
BERT with text.
For LGBM, we simply concatenate the frame representation from the input story blocks to create the input vector.
For BERT, we put the event tuple and the text together as the input.
Table~\ref{tab:history-exp} shows the results. 
While BERT does not benefit from using more contexts, LGBM's performance improves, suggesting the potentials of using a larger context window.
More research is required to understand the effects.



\begin{table}[t]
    \centering \small
    \addtolength{\tabcolsep}{-0.1cm}
    \begin{tabular}{lp{4.4cm}}
        \toprule \hline
        \multicolumn{1}{c}{\textbf{Frame}} & \multicolumn{1}{c}{\textbf{Lexical Units}} \\ \hline
        \multicolumn{2}{c}{Most Important Frames (Out of 50)} \\ \hline
        Kinship & father, mother, son, daughter  \\
        Biological\_urge & tired, sleepy, randy, hungry \\
        Connectors & ribbon, rope, thread, string \\
        Firefighting & fight, battle, control, tackle \\
        Origin & Chinese, American, Vietnamese, origin \\ \midrule
        \multicolumn{2}{c}{Least Important Frames (Out of 50)} \\ \hline
        Proper\_reference &  proper, self\\
        Cause\_to\_start &  spark, generate, arouse, bring about\\
        Friction &  grate, squeal, scrunch, screech \\
        Dominate\_competitor &  dominate, domination, dominant, strongman \\
        State\_continue & remain, stay, rest \\
        \hline \bottomrule
    \end{tabular}
    \addtolength{\tabcolsep}{+0.1cm}
    \caption{The most and least important five frames (from 50 random frames) identified in the ablation study.}
    \label{tab:ablation-exp}
\end{table}

\subsection{Which Semantic Frames Affect the Follow-Up Story More?}
Different frames may contribute differently to the prediction of the follow-up story.
To understand which frame plays a more important role in the story, we conduct an ablation study by investigating the LGBM model on block 150.
We obliterate one frame from the input frame representation and record the performance change, where a higher performance deduction means the frame removed is more important.
A total of 50 frames are selected randomly for the ablation study.
Table~\ref{tab:ablation-exp} shows the top and bottom five frames.
We hypothesize that the more generic frames, such as ``State\_continue'' and ``Proper\_reference,'' might be less important to the follow-up stories, but it will require more research to understand the impacts fully.

\section{Human Evaluation}
\begin{figure*}
    \centering
    \includegraphics[width=\linewidth]{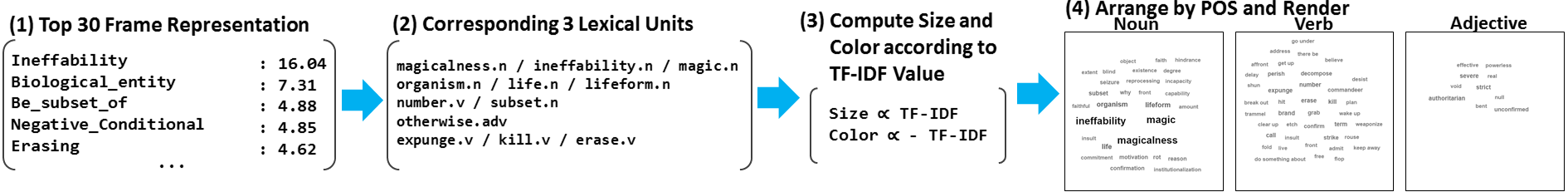}
    \caption{The workflow to visualize the word clouds from frame representation. The top semantic frames are used where each is illustrated by a maximum of three corresponding lexical units. The size and the color of the lexical units are computed according to the TF-IDF value.}
    \label{fig:word-cloud-workflow}
\end{figure*}

We further evaluate the proposed method with humans.
We first visualize the vector of semantic frames into \textbf{word clouds} so that humans can perceive and comprehend it.
We then use online crowd workers to test the
\textit{(i)} representativeness 
and
\textit{(ii)} 
the specificity of the produced word clouds.

\paragraph{Visualizing Semantic Frame Vectors into Word Clouds.}
Figure~\ref{fig:word-cloud-workflow} shows the workflow of generating word clouds based on a frame representation ({\em i.e.,} a TF-IDF vector).
In FrameNet, ``lexical units'' are the terms that can trigger a specific frame.
Compared to showing the name and definition of a frame, lexical units are easier for people to read and comprehend.
Therefore, we use the top 30 frames (ranked by their TF-IDF weights) and
randomly select up to three lexical units for each frame to form a word cloud.
The size and color of the lexical unit is computed according to the frame's TF-IDF weight, where a higher TF-IDF value will result in a larger font and darker color.
Finally, we arrange the lexical units into three word clouds on nouns, verbs, and adjectives using their POS tags.
All the word clouds are generated using \texttt{d3-cloud}~\cite{davies2016d3}.

\subsection{Representativeness}
\label{subsec:rep}
This task evaluates which model can generate the most representative word cloud for a story block.

\paragraph{Task Setups.}
\cy{Clarify story block setup.}
In this Human Intelligence Task (HIT), we show a story block ($n+1$) and two or three [noun, verb, adjective] word clouds ($n+1$) produced by different models based on the previous story block ($n$).
The goal is to measure, from the users' perspective, how much the generated word clouds represent the actual human-written follow-up stories. 
We display the actual next story block ($n+1$) and the word clouds produced by different models based on the latest story block ($n$).
The workers from Amazon Mechanical Turk (MTurk) are asked to read the story and select the word cloud that better represents the story block.
In the worker interface, we set up a 3-minutes lock for submission and a reach-to-the-bottom lock for the story panel to make sure the workers read the story.
Nine different workers are recruited for each task\footnote{Four built-in worker qualifications are used: HIT Approval Rate ($\geq $98\%), Number of Approved HITs ($\geq 3000$), Locale (US Only), and Adult Content Qualification.}.
We empirically estimate the working time to be less than 6 minutes per HIT and set the price to \$0.99/HIT (hourly wage = \$10).


We choose block size 150 to compare two models: LGBM with frame representation and BERT with text. 
Ground-truth word clouds are also added to some of the HITs to check the validity of the task.
A total of 150 instances are randomly selected from Boocorpus testing set. 
For each instance, the foregoing story block is feed into LGBM and BERT to predict the frame representation of the follow-up story block. 
Out of 150 instances, 50 instances are conducted with ground truth, where a total of three word clouds are shown. 
Another 100 instances are used for comparing LGBM against BERT directly.


\paragraph{Results.}
Over the 50 HITs where ground truth is included, (ground truth, LGBM, BERT) wins (32, 15, 16) HITs, respectively (ties exist.)
Nine assignments are recruited from 9 workers for each HIT.
Regarding to the assignment voting, (ground truth, LGBM, BERT) gets (199, 131, 120) votes, respectively.
The result suggests that humans can correctly perceive the word clouds' conceptual meaning as the ground truth is rated the best.

Over the 100 HITs where LGBM and BERT are compared directly, (LGBM, BERT) wins (59, 41) HITs.
Regarding the assignment voting, (LGBM, BERT) gets (472, 428) votes, respectively.
The result shows that LGBM is better than BERT in a block size of 150, which aligns with our automatic evaluation results using cosine similarity (see Section~\ref{sec:results}.)

\subsection{Specificity}
This task evaluates whether using the proposed word cloud to represent a story block is specific enough for humans to distinguish the correct story from the distractor.

\paragraph{Task Setups.}
\cy{Clarify the setup.}
In this HIT, we show two story blocks ($n$) and one set of [noun, verb, adjective] word clouds ($n$).
Note that the current story block ($n$) and its ground-truth word cloud ($n$) are used to examine if 
humans can correctly perceive the semantic information from word cloud visualization.
One story block is the answer that is referred to by the word clouds and the other one is a distractor.
Workers are asked to read the two story blocks and select the story block that is referred to by the word clouds.
Nine different workers are recruited for each HIT. 
We use the same worker interface design and built-in worker qualifications as that of Section~\ref{subsec:rep}.
A HIT takes estimatedly 2.33 minutes and is priced at \$0.38.



We choose block size 20 and use the ground-truth word clouds for this experiment. 
Fifty instances from 50 different books are randomly selected from Bookcorpus testing set. 
We also randomly select a 20-sentences story block from a different book as the distractor.



\paragraph{Results.}

Of the 450 assignments, 63.8\% of the answers were correct.
When aggregating the assignments using majority voting, 74\% of 50 HITs were answered correctly.
We thus believe that it is reasonably specific for humans to 
represent a story block using the proposed word clouds.

\section{Conclusion}

This paper proposes a \task task that aims to forecast the semantic frames in the next 10, 100, or even 1,000 sentences of a story. 
A long story is formulated as a sequence of story blocks that contain a fixed number of sentences.
We further introduce a frame representation that can encode a story block into a fixed-length TF-IDF vector over semantic frames.
Experiments on both the Bookcorpus dataset and CODA-19 dataset show that the proposed frame representation helps \task in large story blocks. 
By visualizing the frame representation as word clouds, we also show that it is comprehensible, representative, and specific to humans. 
In the future, we will introduce the frame representation into story generation models to ensure coherence when generating long stories. 
We will also explore the possibility of supporting writers to develop the next part of their stories by generating semantic frames as clues using \task.


\section*{Acknowledgments}
We would like to thanks the Huck Institutes of the Life Sciences' Coronavirus Research Seed Fund (CRSF) and the College of IST COVID-19 Seed Fund at Penn State University who support the construction of CODA-19.
We also thank Tiffany Knearem for the feedback for designing word cloud visualization and workers who participated the human evaluation study.

\bibliography{anthology,custom}
\bibliographystyle{acl_natbib}



\end{document}